\documentclass[a4paper,conference]{IEEEtran}

\usepackage[english]{babel}
\usepackage{hyperref}
\usepackage{booktabs}
\usepackage{multirow}
\usepackage{amsmath}
\usepackage{balance}
\usepackage[pdftex]{graphicx}
\usepackage{amsfonts}

\begin{document}
\title{Boosting Camouflaged Object Detection with Dual-Task Interactive Transformer}

\author{\IEEEauthorblockN{Zhengyi Liu*}
\IEEEauthorblockA{Email: liuzywen@ahu.edu.cn}
\and
\IEEEauthorblockN{Zhili Zhang}
\IEEEauthorblockA{Email: 528419003@qq.com}
\and
\IEEEauthorblockN{Wei Wu}
\IEEEauthorblockA{Email: 2640947588@qq.com}}
\maketitle
\begin{abstract}
Camouflaged object detection intends to discover the concealed objects hidden in the surroundings. Existing methods follow the bio-inspired framework, which first locates the object and second refines the boundary. We argue that the discovery of camouflaged objects depends on the recurrent search for the object and the boundary. The recurrent processing makes the human tired and helpless, but it is just the advantage of the transformer with global search ability. Therefore, a dual-task interactive transformer is proposed to detect both accurate position of the camouflaged object and its detailed boundary. The boundary feature is considered as Query to improve the camouflaged object detection, and meanwhile the object feature is considered as Query to improve the boundary detection. The camouflaged object detection and the boundary detection are fully interacted by multi-head self-attention. Besides, to obtain the initial object feature and boundary feature, transformer-based backbones are adopted to extract the foreground and background. The foreground is just object, while foreground minus background is considered as boundary. Here, the boundary feature can be obtained from blurry boundary region of the foreground and background. Supervised by the object, the background and the boundary ground truth, the proposed model achieves state-of-the-art performance in public datasets.
\href{https://github.com/liuzywen/COD}{https://github.com/liuzywen/COD}
\end{abstract}

\section{Introduction}
In the world of animals, prey often evolves to have a similar
appearance to their surroundings and thus deceives the predators. It is called camouflage which also happens in the human (e.g., military camouflage clothes).
Camouflaged object detection (COD) is challenging due to the minor difference between camouflaged objects and their surroundings in color, texture, brightness, etc.
Nevertheless, COD deserves well exploration because it has important scientific and  practical application value in medical diagnosis \cite{dong2021polyp,kim2021uacanet,fan2020inf}, industrial defect detection \cite{le2020learning}, agricultural pest identification \cite{dai2017convolutional,lev2004plant} and so on.

Traditional low-level hand-crafted features (e.g., edge, brightness,
color, gradient, texture) \cite{tankus2001convexity, bhajantri2006camouflage, feng2015camouflage, xue2016camouflage, pike2018quantifying, li2018fusion} seem to be not always valid because well-performed camouflage is skilled in breaking the low-level features.
Convolution Neural Network (CNN) has powerful representative
ability to extract both high-level semantic features and low-level spatial information.
It is widely adopted as backbone to establish multi-level and multi-scale features.
Based on CNN backbone, SINet \cite{fan2020camouflaged}, SINet-V2 \cite{fan2021concealed}, 
PFNet \cite{mei2021camouflaged} 
and BGNet \cite{xu2021boundary} mimicked two-stage detection process of human visual mechanism.
They first locate rough areas and then accurately segment camouflaged objects.
TANet \cite{ren2021deep}, Depth-guidedCOD \cite{zhang2021depth}, JCOD \cite{li2021uncertainty} and CANet \cite{liu2021confidence} identified camouflaged object by amplifying subtle texture difference, estimating depth information, jointly training salient object detection task and measuring confidence, respectively.
Recently, transformer has achieved revolutionary progress in various vision tasks. 
TransformerCOD \cite{mao2021transformer} adopts Swin Transformer as backbone, and introduces a generator and a discriminator to estimate pixel-wise confidence map and achieve difficulty-aware learning.
UGTR \cite{yang2021uncertainty} uses a probabilistic representational  model  to
reason under uncertainties by transformer.
Inspired by them, we adopt transformer-based backbone to capture global contextual information to establish long-range dependency, and meanwhile retain multi-stage and multi-scale characteristic which is adaptive to the objects with different sizes like CNN.

Furthermore, we find that boundary features play a crucial role in dense prediction tasks.
Existing methods adopted boundary-sensitive loss functions (e.g., NLDF \cite{luo2017non} and BASNet \cite{qin2019basnet,qin2021boundary}), or introduced boundary supervision (e.g., EGNet \cite{zhao2019egnet}, PAGE-Net \cite{wang2019salient} and ERRNet \cite{ji2021fast}), or adopted interactive CNN structure (e.g., MGL \cite{zhai2021mutual} and BGNet \cite{xu2021boundary}) to boost boundary detection and object detection.

In this paper, we propose a dual-task interactive transformer to promote the interaction between boundary detection and COD task. The boundary feature can provide the outline details of camouflage objects, and meanwhile camouflage object feature gives the guidance to boundary refinement in position cue.
Different from two-stage bio-inspired location and refinement strategy (e.g., SINet \cite{fan2020camouflaged}), we design the network model to search object and boundary in parallel by cross multi-head self-attention, attempting to  exceed the human intelligence, rather than mimic human vision.

In addition, we design a boundary feature extraction method. It uses two backbones to extract foreground and background features, respectively. Then the minus operation is applied on foreground and background features, generating boundary features. Uncertain regions in foreground and background features are reassigned to sharp boundary information.

Finally, we conduct experiments on public benchmark datasets and verify the effectiveness of our design.

Our main contributions are highlighted as follows:
\begin{itemize}
\item Different from bio-inspired framework, we use transformer with great global search ability to interactively achieve camouflaged object detection task and boundary detection task, improving the performance of COD.
\item Different from existing boundary extraction from one backbone, we use two backbones to extract the foreground and background features, and then view foreground minus background as boundary.  Certain foreground and background and uncertain parts show the different distribution. The designed boundary extraction strategy converts uncertain region of foreground and background to  boundary feature.
\item The proposed model is supervised by object, background and boundary ground truth. It achieves an impressive improvement in COD task.
\end{itemize}
\section{Proposed Method}
\begin{figure*}[htp!]
\centering
\includegraphics[width=\textwidth]{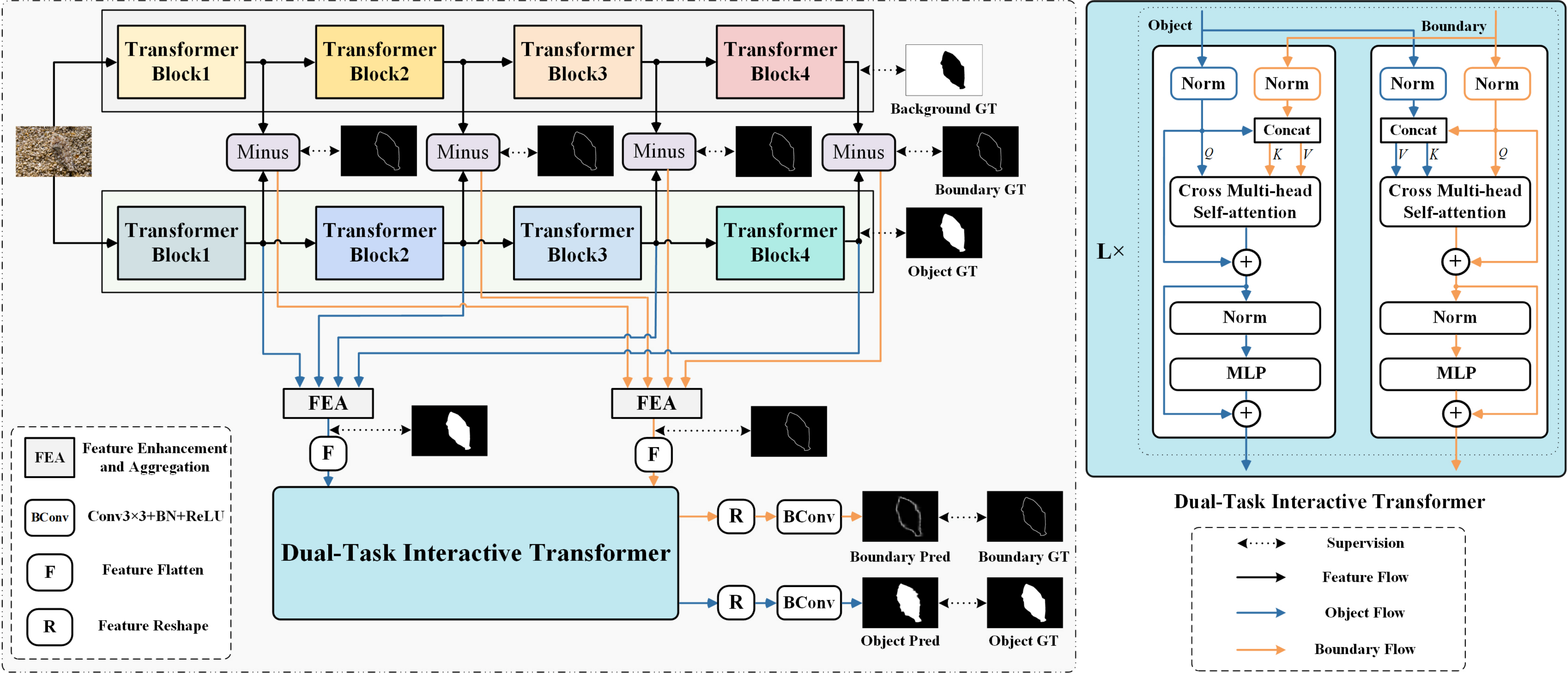}
\caption{The proposed model and dual-task interactive transformer.}
\label{fig:main}
\end{figure*}
\subsection{Motivation}
Camouflaged object detection (COD) aims to segment camouflaged objects hiding in the environment.
The key challenge lies in the fact that camouflaged objects are similar with their surroundings in color, texture, brightness or other patterns.
Therefore, traditional contrast-based methods no longer work.
But we find that although camouflaged object can hide itself in the background, it inevitable has its boundary.
Therefore, boundary detection (BD) is important for COD task.
Once object is found, we want to further identify the boundary, and then maybe the combination of object and  boundary  is judged as error, we further search the other region to finally find the accurate object and its boundary. Its process can be performed by computer with parallel and fast computation ability. Therefore, different from
existing bio-inspired methods (e.g., SINet \cite{fan2020camouflaged}, PFNet \cite{mei2021camouflaged}) which find camouflaged objects by search-first and identification-second processes,
we propose a dual-task interactive transformer to  detect the camouflaged objects and boundary in parallel, improving the performance of COD task.
Our work first solves the COD task using parallel and global search ability of transformer, and attempts to exceed human rather than mimic human vision system.
\subsection{Overview of Our Proposed Model for COD}
The proposed model is an encoder-decoder framework, as shown in Fig.\ref{fig:main}.
The encoder is responsible for extracting two kinds of features which are object feature and boundary feature.
The decoder is used to achieve two kinds of tasks in  parallel, which are camouflaged object detection (COD) task and  boundary detection (BD) task.
The encoder consists of two backbones for foreground and background extraction, boundary generation by minus operation, enhancement and aggregation of object and boundary feature.
The decoder is composed of dual-task interactive transformer and prediction module.
\subsection{Two Backbones for Foreground and Background Extraction}
Recently, many transformer-based backbones are proposed, for example, ViT \cite{dosovitskiy2020image}, CvT \cite{wu2021cvt}, T2T \cite{yuan2021tokens}, Twins \cite{chu2021twins}, Swin Transformer \cite{liu2021swin}, PVT \cite{wang2021pyramid,wang2021pvtv2}, SegFormer \cite{xie2021segformer}.
In these backbones, SegFormer has both global and local attributions owing to the multi-head self-attention and overlapping patch merging. It uses depth-wise convolutions to reduce the number of parameters, and adopts sequence reduction process to reduce the computational complexity.
Thus, it achieves state-of-the-art performance in semantic segmentation task.
Due to its excellent performance,
we adopt two SegFormers as the encoder to extract foreground and background  features from input image. Considering the complexity and efficiency, the MiT-B5 version \cite{xie2021segformer} is adopted.
Each SegFormer first splits input image into overlapping patches, and these patches are then fed into the positional-encoding-free multi-stage feature transformation, wherein multi-head self-attention and feed-forward network model the relation between tokens.
As the  depth of network increases, the number of tokens gradually reduces
to generate a hierarchical representation of foreground stream and background stream, which can be denoted as object feature $\{f_i^o\}_{i=1}^4$ and background feature $\{f_i^b\}_{i=1}^4$. 

In order to make two backbones extract more information about foreground and background, we use camouflaged object ground truth ($GT$) to supervise foreground stream, and use 1-$GT$ to supervise background stream. Here, loss function adopts pixel position aware loss \cite{wei2020f3net}.
\subsection{Boundary Generation by Minus Operation}
Two backbones are responsible for camouflaged object foreground detection and background detection, respectively.
In general, extracted features have certain foreground or certain background part and uncertain part. Suppose foreground and background features are within [0,1], difference of certain parts will tend to -1 or 1, while difference of uncertain parts tends to 0. Certain foreground minus certain background and that of uncertain parts show the different distribution. Therefore, we use minus operation to reassign uncertain part to the right position of boundary. That is to say, blurry boundary region of foreground and background is converted to sharp boundary by minus operation between foreground and background feature.
It can be described as:
\begin{equation}
\begin{aligned}
f^e_i=BConv(BConv(f^o_i)-BConv(f^b_i))
\end{aligned}
\end{equation}
where $f^e_i$ is boundary feature, in which $i$ represents level number of the feature, $BConv(\cdot)$ means convolution operation, followed by a batch normalization and a ReLU activation function. All the boundary features $\{f_i^e\}_{i=1}^4$ are supervised by boundary ground truth which is obtained  from camouflaged object ground truth (GT) by Canny edge detector \cite{canny1986computational}. Here, loss function adopts cross entropy loss.

\subsection{Feature Enhancement and Aggregation}
After feature extraction of two backbones and boundary generation, we obtain two kinds of useful features which are object features $f^o_i(i=1,\cdots,4)$ and boundary features $f^e_i(i=1,\cdots,4)$.
These hierarchical features with different scales and receptive fields
are important for COD task, for they are adaptive to detect objects with different sizes.
Next, all the object features and all the boundary features need to be enhanced and aggregated, respectively.
Inspired by cascaded fuse module proposed in \cite{dong2021polyp} and cascaded decoder proposed in \cite{fan2020bbs}, we first enhance the features in different layers and then aggregate them progressively.
\begin{figure}[htp!]
\centering
\includegraphics[width=\linewidth]{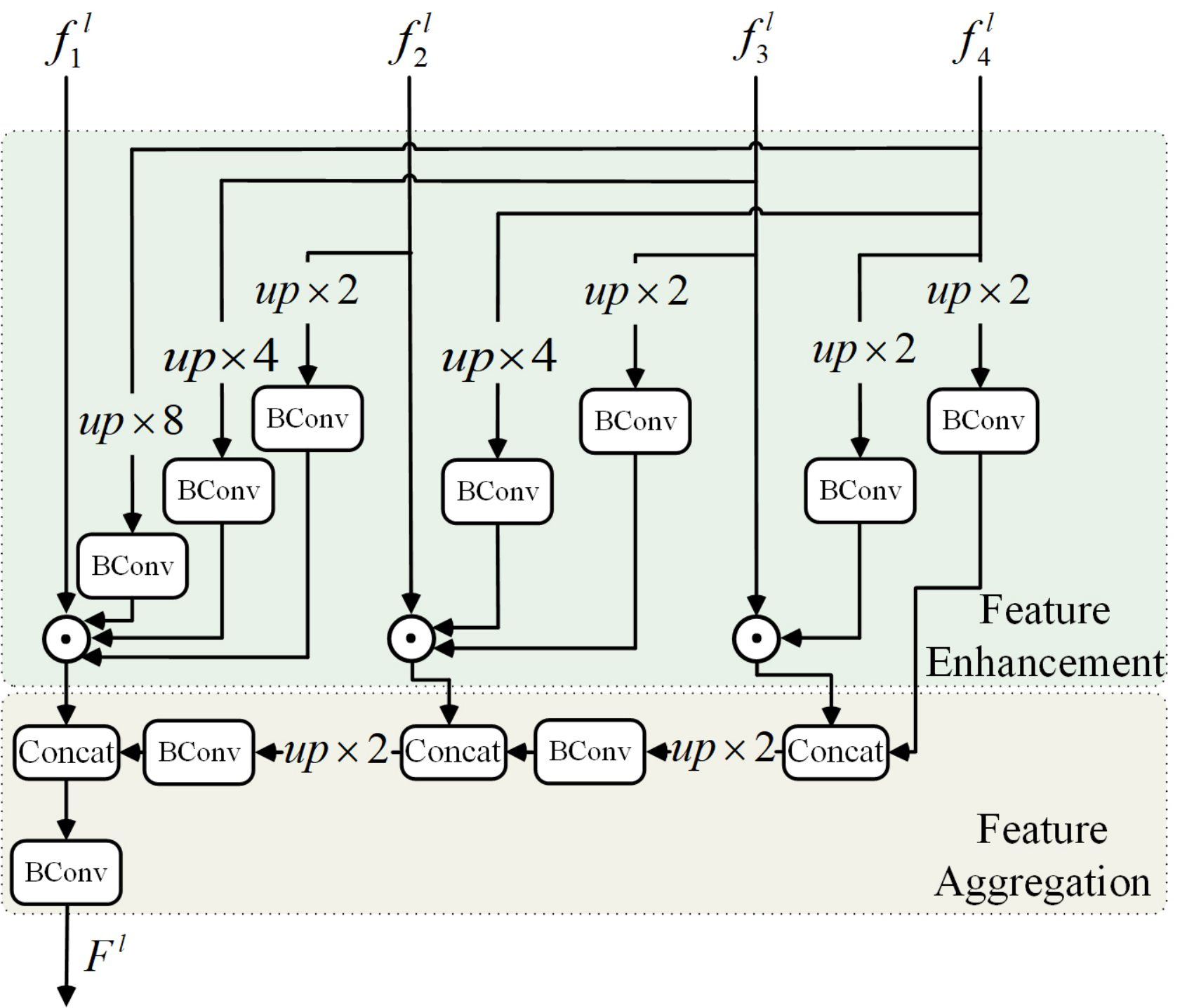} 
\caption{Feature enhancement and aggregation.}
\label{fig:FEAM}
\end{figure}

Specifically, as shown in Fig.\ref{fig:FEAM}, in enhancement part, features in the higher layer are upsampled to the same size with the feature in the current layer, and then they are combined by multiplication operation.
It can be described as:
\begin{equation}
\begin{aligned}
g^l_i= \left\{\begin{matrix}
	f^l_i\odot \prod^4_{j=i+1}{BConv(Up(f^l_j))},&i=1,2,3 \\
	f^l_i,&i=4
	\end{matrix}\right.
\end{aligned}
\end{equation}
where $l\in \{o,e\}$ represents object feature or boundary feature, $\odot(\cdot)$ and $\prod(\cdot)$ denote the Hadamard product or successive Hadamard product, and $Up(\cdot)$ denotes upsampling operation.

In aggregation part, the enhanced features are progressively concatenated and upsampled to generate aggregated feature.
It can be described as:
\begin{equation}
\begin{aligned}
q^l_{i}= \left\{\begin{matrix}
	Concat_{channel}(g^l_i,BConv(Up(q^l_{i+1}))),&i=1,2,3 \\
	g^l_i,&i=4
	\end{matrix}\right.
\end{aligned}
\end{equation}
where $Concat_{channel}(\cdot)$ denotes concatenation operation along the channel dimension.

At last, the progressive aggregation generates object feature $F^o$ or boundary feature $F^e$ by the following formula:
\begin{equation}
\begin{aligned}
F^l=BConv(q^l_{1})
\end{aligned}
\end{equation}
where $F^l$ represents object feature or boundary feature which will be fed into next dual-task interactive transformer for interaction.
\subsection{Dual-task Interactive Transformer}
Our dual-task interactive transformer processes camouflaged object detection task and boundary detection task in parallel, which employs two symmetric branches, i.e., BD branch and COD branch.

Before COD and BD branch, object feature and boundary feature $F^{l}\in \mathbb{R}^{H\times W\times C}(l\in\{o,e\})$ are first split into a sequence of flattened 2D patches $\{F^{l}_p\in \mathbb{R}^{P^{2}\times C}|p=1,\cdots, N\}$, where $N=\frac{HW}{P^2}$ is the number of patches, and $P\times P$ is the resolution of each patch.
Next, each patch $F^{l}_p$ is then mapped into a latent $D$-dimensional embedding space by a trainable linear projection layer. Furthermore, we learn specific position embeddings which are added to the patch embedding to retain positional information. The process can be described as:
\begin{equation}
\begin{aligned}
z^l_0=[F^{l}_1E^{l};F^{l}_2E^{l};\cdots;F^{l}_NE^{l}]+E_{pos}^{l}
\end{aligned}
\end{equation}
where $E^{l}\in \mathbb{R}^{(P^2\times C)\times D}$ is patch embedding projection, and $E^{l}_{pos}\in \mathbb{R}^{N\times D}$ denotes the position embedding.

The remaining architecture stacks $L$ dual-task interactive transformer layers.
Each dual-task interactive transformer layer contains a cross multi-head self-attention (CMSA) and multi-layer perceptron (MLP) sublayer. Besides, residual connections are
applied after CMSA and MLP.
The process can be described as:\\
\begin{equation}
\left\{\begin{array}{l}
{z^{l}_i}^{\prime}=CMSA\left(z^{l}_{i-1},z^{\bar l}_{i-1}\right)+z^{l}_{i-1} \\
z^{l}_i=MLP\left(LN\left({z^{l}_i}^{\prime}\right)\right)+{z^{l}_i}^{\prime}
\end{array} \quad i=1, \cdots L\right.
\end{equation}
where $LN(\cdot)$ represents layer normalization.
$\bar l$ denotes the other variant in the set $\{o,e\}$ except $l$.

In order to make COD task and BD task interact with each other, and further improve respective performance, the CMSA absorbs features from one task that contributes to the other task.
Specifically, the features in one task serve as a query that interacts with the features from the other task through attention.
For COD branch, we use the features $z^o_{i}$ after layer normalization $LN(z^o_{i})$ as query ($Q$), and concatenate them with the features from BD branch $LN(z^e_{i})$ to serve as key ($K$) and value ($V$).
Similarly, for BD branch, we use the features $z^e_{i}$ after layer normalization $LN(z^e_{i})$ as  query ($Q$), and concatenate them with the features from COD branch $LN(z^o_{i})$ to serve as key ($K$) and value ($V$).
Then, the interaction between two branches can be performed by CMSA, and it is defined as:
\begin{equation}
\begin{aligned}
CMSA\left(z^{l}_{i},z^{\bar l}_{i}\right)=softmax(\frac{QK^T}{\sqrt{D/h}})V
\end{aligned}
\end{equation}
where  $D$ is embedding dimension and $h$ is the number of heads, $Q$, $K$ and $V$ are defined as:
\begin{equation}
\begin{aligned}
Q=LN(z^{l}_{i})W_Q
\end{aligned}
\end{equation}
\begin{equation}
\begin{aligned}
K=Concat_{patch}(LN(z^{l}_{i}),LN(z^{\bar l}_{i}))W_K
\end{aligned}
\end{equation}
\begin{equation}
\begin{aligned}
V=Concat_{patch}(LN(z^{l}_{i}),LN(z^{\bar l}_{i}))W_V
\end{aligned}
\end{equation}
where $Concat_{patch}(\cdot)$ means patch concatenation, $W_Q$, $W_K$, $W_V$ $\in \mathbb{R}^{D\times D}$ are projection matrix of full connection layer.
\subsection{Prediction Module}
After multiple dual-task interactive transformer layers, object feature and boundary feature are obtained from $z^{l}_L$, they are further reshaped from patch sequence to feature. Finally, predicted object and boundary results are generated by smooth operation. The process can be described as:
\begin{equation}
\begin{aligned}
S^l=Sig(Up(Conv(BConv(Reshape(z^{l}_L)))))
\end{aligned}
\end{equation}
where $S^o$ and $S^e$ represent camouflaged object prediction map and boundary prediction map. The operation $Reshape(\cdot)$ reshapes patch sequences  with $\frac{HW}{P^2}\times D$ to feature with $\frac{H}{P}\times \frac{H}{P}\times D$, $BConv(\cdot)$, $Conv(\cdot)$ and $Up(\cdot)$ operation further change the channel and resolution, and last $Sig(\cdot)$ performs  Sigmoid activation function.

The predicted camouflaged map and boundary map are supervised by object ground truth and boundary ground truth. Here, loss function for camouflaged map adopts pixel position aware loss, and loss function for boundary map adopts cross entropy loss.

\section{Experiments}
\subsection{Experimental Setup}
\subsubsection{\textit{Datasets}}
We perform extensive experiments on the following
public benchmarks:
\begin{itemize}
\item \underline{CAMO} \cite{le2019anabranch} includes 1,250 camouflaged images divided into 8 categories, where 1,000 camouflaged images are for training, and the remaining 250 images are for testing.
\item \underline{COD10K} \cite{fan2021concealed,fan2020camouflaged} contains 5,066 camouflaged images downloaded from multiple photography websites, covering 5 super-classes and 69 sub-classes. It released 3,040 camouflaged images for training and 2,026 images for testing.
\item \underline{NC4K} \cite{lv2021simultaneously} provides the largest testing dataset with 4,121 images for effective model evaluation.
\end{itemize}
The same training datasets are adopted as SINet \cite{fan2021concealed}.
Note that since CHAMELEON \cite{skurowski2018animal} dataset has only 76 images downloaded from the Internet, we don't use it due to possible bias.
\subsubsection{\textit{Evaluation Metrics}}
We adopt five widely used evaluation metrics to evaluate the performance of COD
models, including precision-recall (PR) curve, S-measure ($S_\alpha$), mean E-measure ($E_\phi$), weighted F-measure ($F^\omega_\beta$) and Mean Absolute Error(MAE).
Specifically, the PR curve plots precision and recall values by setting a series of thresholds on the camouflaged maps to get the binary masks and further comparing them with the ground truth maps.
The S-measure can evaluate both region-aware and object-aware structural similarity between camouflaged map and ground truth.
The E-measure simultaneously captures global statistics and local pixel matching information.
The F-measure is the weighted harmonic mean of precision and recall, which can evaluate the overall performance.
The MAE measures the average of the per-pixel absolute difference between the camouflaged maps and the ground truth maps.

\subsubsection{\textit{Implementation Details}}
We implement the proposed model with PyTorch framework and use a RTX 3090 GPU to accelerate the calculations. During training, for data augmentation, we use horizontal flip, random crop and random rotation. The backbone parameters of Segformer \cite{xie2021segformer} are initialized with the corresponding models pre-trained on ImageNet dataset, dual-task interactive transformer is initialized with pre-trained parameters of ViT \cite{dosovitskiy2020image} and the rest parameters are randomly initialized. Note that hyper-parameters in dual-task interactive transformer are set as: $L$=6, $D$=768, $h$=12, $P$=2.
We use the Adam optimizer, which is widely used in transformer networks. The learning rate is set to 6e-5.
Further, we resize the input images to 256$\times$256 with a batch size of 4 for 100 epochs, which takes nearly 20 hours.
During testing, we resized each image to 256$\times$256 and then feed it to model for predicting camouflaged maps without any post-processing. 

\subsection{Comparison with State-of-the-Arts}
We compare the proposed model with state-of-the-art
methods, including SINet \cite{fan2020camouflaged}, ERRNet \cite{ji2021fast}, C2F-Net \cite{sun2021context}, 
Rank-Net \cite{lv2021simultaneously}, MGL \cite{zhai2021mutual}, JCOD \cite{li2021uncertainty},
PFNet \cite{mei2021camouflaged},  SINet-V2 \cite{fan2021concealed} and UGTR \cite{yang2021uncertainty}. For fair comparison, all the camouflaged
maps of above approaches are provided by the authors.
\subsubsection{\textit{Quantitative Evaluation}}
\begin{table*}[!htp]
    \centering
    \setlength{\tabcolsep}{1.0mm}{
    \caption{S-measure, E-measure, F-measure, MAE comparisons with different COD models.}
    \label{tab:comparisonCOD}
    \resizebox{0.6\textwidth}{!}
{
\begin{tabular}{c|c|cccc|cccc|cccc}
    \hline\toprule
   \multirow{2}{*}{\centering Model} & \multirow{2}{*}{\centering Source} & \multicolumn{4}{c|}{\centering CAMO} & \multicolumn{4}{c|}{\centering COD10K} & \multicolumn{4}{c}{\centering NC4K}\\
     &&$S_\alpha \uparrow$ & $E_\phi \uparrow$ &$F^\omega_\beta\uparrow$ & MAE$\downarrow$
     &$S_\alpha \uparrow$ & $E_\phi \uparrow$ &$F^\omega_\beta\uparrow$ & MAE$\downarrow$
     &$S_\alpha \uparrow$ & $E_\phi \uparrow$ &$F^\omega_\beta\uparrow$ & MAE$\downarrow$ \\
    \hline
    SINet & CVPR20
        &.751   &.771   & .606  &.100
        &.771   &.806   & .551  &.051
        &.808   &.871   & .723  &.058 \\
    ERRNet & PR21
        &.778   &.842   & .679  &.085
        &.786   &.867   & .630  &.043
        &.827   &.887   & .737  &.054 \\
    SINet-V2 & PAMI21
        &.822   &.882   & .743  &.070
        &.815   &.887   & .680  &.037
        &.847   &.903   & .770  &.048 \\
    C2F-Net & IJCAI21
        &.796   &.854   & .719  &.080
        &.813   &.890   & .686  &.036
        &.838   &.897   & .762  &.049 \\
    Rank-Net & CVPR21
        &.787   &.838   & .696  &.080
        &.804   &.880   & .673  &.037
        &.840   &.895   & .766  &.048 \\
    MGL & CVPR21
        &.775   &.726   & .673  &.088
        &.814   &.852   & .667  &.035
        &.832   &.867   & .740  &.052 \\
    JCOD & CVPR21
        &.800   &.859   & .728  &.073
        &.809   &.884   & .684  &.035
        &.842   &.898   & .771  &.047 \\
    PFNet & CVPR21
        &.782   &.841   & .695  &.085
        &.800   &.877   & .660  &.040
        &.829   &.887   & .745  &.053 \\
    UGTR & ICCV21
        &.784   &.823   & .686  &.086
        &.818   &.853   & .667  &.035
        &.839   &.874   & .747  &.052 \\
    Ours & ICPR22
        &\textbf{.857}&\textbf{.916}&\textbf{.796}&\textbf{.050}
        &\textbf{.824}&\textbf{.896}&\textbf{.695}&\textbf{.034}
        &\textbf{.863}&\textbf{.917}&\textbf{.792}&\textbf{.041} \\
    \bottomrule
    \hline
\end{tabular}}}
\end{table*}

\begin{figure*}[!htp]
\centering
\begin{tabular}{ccc}
\includegraphics[width = 0.33\textwidth]{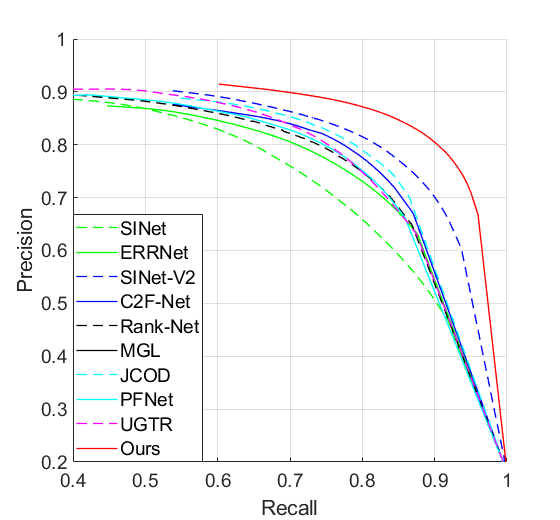}&\includegraphics[width = 0.33\textwidth]{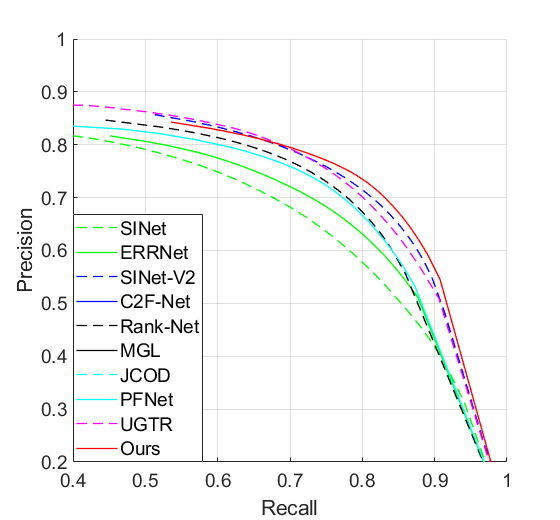}&\includegraphics[width = 0.33\textwidth]{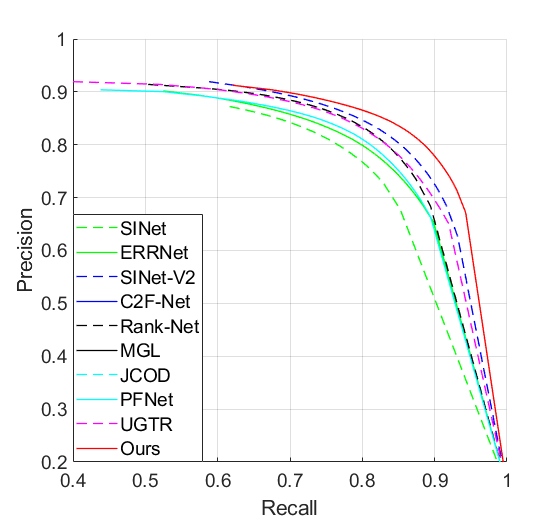}\\
(a) CAMO dataset&(b) COD10K dataset&(c) NC4K dataset\\
\end{tabular}
\caption{P-R curves comparisons of different COD models.}
\label{fig:PRComparison}
\end{figure*}

Table.~\ref{tab:comparisonCOD} shows the detailed quantitative results on public datasets.
As can be seen clearly that our method outperforms state-of-the-art models in terms of all four evaluation metrics. $S_\alpha$, $E_\phi$, $F^\omega_\beta$ and MAE are improved 0.020, 0.029, 0.030 and 0.010 on average compared the second best SINet-V2.
We also present the precision-recall curves of public
datasets in Fig.\ref{fig:PRComparison}. As can be seen, the curves of our model are consistently higher than the others. It also verifies the effectiveness of proposed model.
\subsubsection{\textit{Qualitative Comparison}}
\begin{figure}[htp!]
\centering
\includegraphics[width=\linewidth]{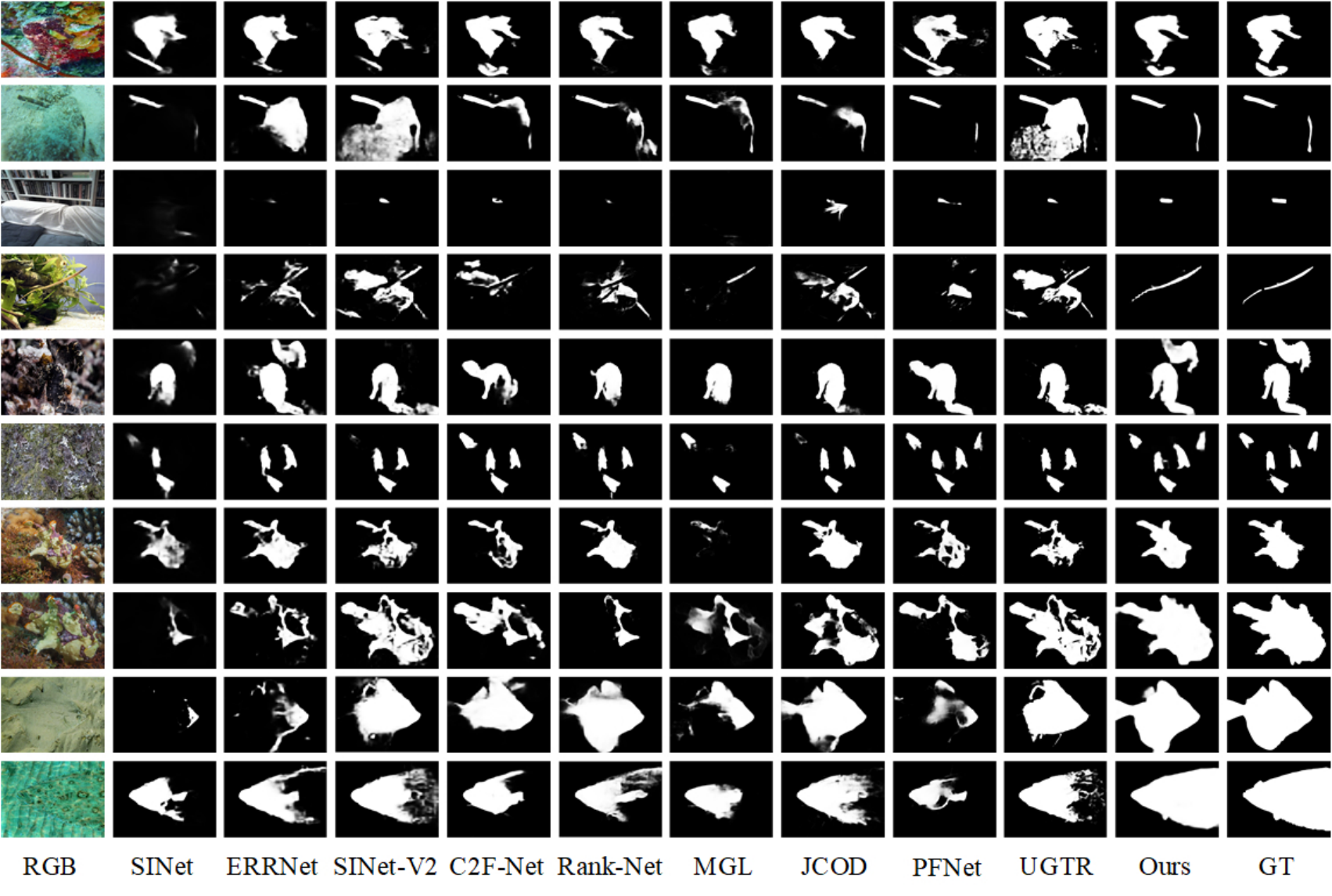} 
\caption{Visual comparison with different COD models in some challenging cases, including occluded objects, small, multiple and big targets, low contrast backgrounds.}
\label{fig:visual}
\end{figure}

Fig.\ref{fig:visual} shows visual comparison results of our model with some state-of-the-art models. It can be intuitively observed that our method successfully segments the position of camouflage objects with accurate boundary under various complex scenes, such as  occluded objects($1^{st}_{}$-$2^{nd}_{}$ rows), small objects($3^{rd}_{}$-$4^{th}_{}$ rows),  multiple objects($5^{th}_{}$-$6^{th}_{}$ rows), big objects($7^{th}_{}$-$8^{th}_{}$ rows) and low-contrast objects($9^{th}_{}$-$10^{th}_{}$ rows).
In the $1^{st}_{}$-$2^{nd}_{}$ rows, our model can successfully segment the object by removing the interference of occlusion.
In the $3^{rd}_{}$-$4^{th}_{}$ rows, our model can be adaptive to small object.
In the $5^{th}_{}$-$6^{th}_{}$ rows, multiple objects are all detected with our model, while some other models have detection loss in the numbers.
In the $7^{th}_{}$-$8^{th}_{}$ rows, our model greatly retains object integrity.
In the $9^{th}_{}$-$10^{th}_{}$ rows, camouflaged objects can be detected from extremely similar surroundings.
The performances in these challenging cases verify the effectiveness of our model.

\subsection{Ablation Study}
\subsubsection{\textit{Effectiveness of Dual-Task Interactive Transformer}}
Table.~\ref{tab:DTITAblation} shows the effectiveness of dual-task interactive transformer (DTIT). In ``EarlyFuse", object feature and boundary feature are directly combined and then fed into a transformer with no interaction.
In ``LateFuse", object feature and boundary feature are fed into two transformers with no interaction, and last combined.
In our proposed model, object feature and boundary feature are fed into two transformers with interaction, i.e., dual-task interactive transformer.
From the comparison, we find that our model uses cross attention to achieve the optimal fusion result, which is represented by the improvement of all evaluation metrics.

\begin{table}[!htp]
    \centering
    \setlength{\tabcolsep}{1.0mm}{
    \caption{Ablation experiments of  Dual-Task Interactive Transformer. The best result is in bold.}
    \label{tab:DTITAblation}

  \resizebox{0.5\textwidth}{!}
{
\begin{tabular}{c|cccc|cccc|cccc}
    \hline\toprule
   \multirow{2}{*}{\centering Variant} & \multicolumn{4}{c|}{\centering CAMO} & \multicolumn{4}{c|}{\centering COD10K} & \multicolumn{4}{c}{\centering NC4K}\\
     &$S_\alpha \uparrow$ & $E_\phi \uparrow$ &$F^\omega_\beta\uparrow$ & MAE$\downarrow$
     &$S_\alpha \uparrow$ & $E_\phi \uparrow$ &$F^\omega_\beta\uparrow$ & MAE$\downarrow$
     &$S_\alpha \uparrow$ & $E_\phi \uparrow$ &$F^\omega_\beta\uparrow$ & MAE$\downarrow$ \\
    \hline
    EarlyFuse
        &.849   &.906   & .778  &.054
        &.814   &.881   & .675  &.037
        &.860   &.912   & .784  &.043 \\
    LateFuse
        &.823   &.868   & .736  &.064
        &.802   &.868   & .653  &.038
        &.845   &.895   & .760  &.046 \\
    Ours
        &\textbf{.857}&\textbf{.916}&\textbf{.796}&\textbf{.050}
        &\textbf{.824}&\textbf{.896}&\textbf{.695}&\textbf{.034}
        &\textbf{.863}&\textbf{.917}&\textbf{.792}&\textbf{.041} \\
    \bottomrule
    \hline
\end{tabular}}
}
\end{table}

\begin{figure}[htp!]
\centering
\includegraphics[width=\linewidth]{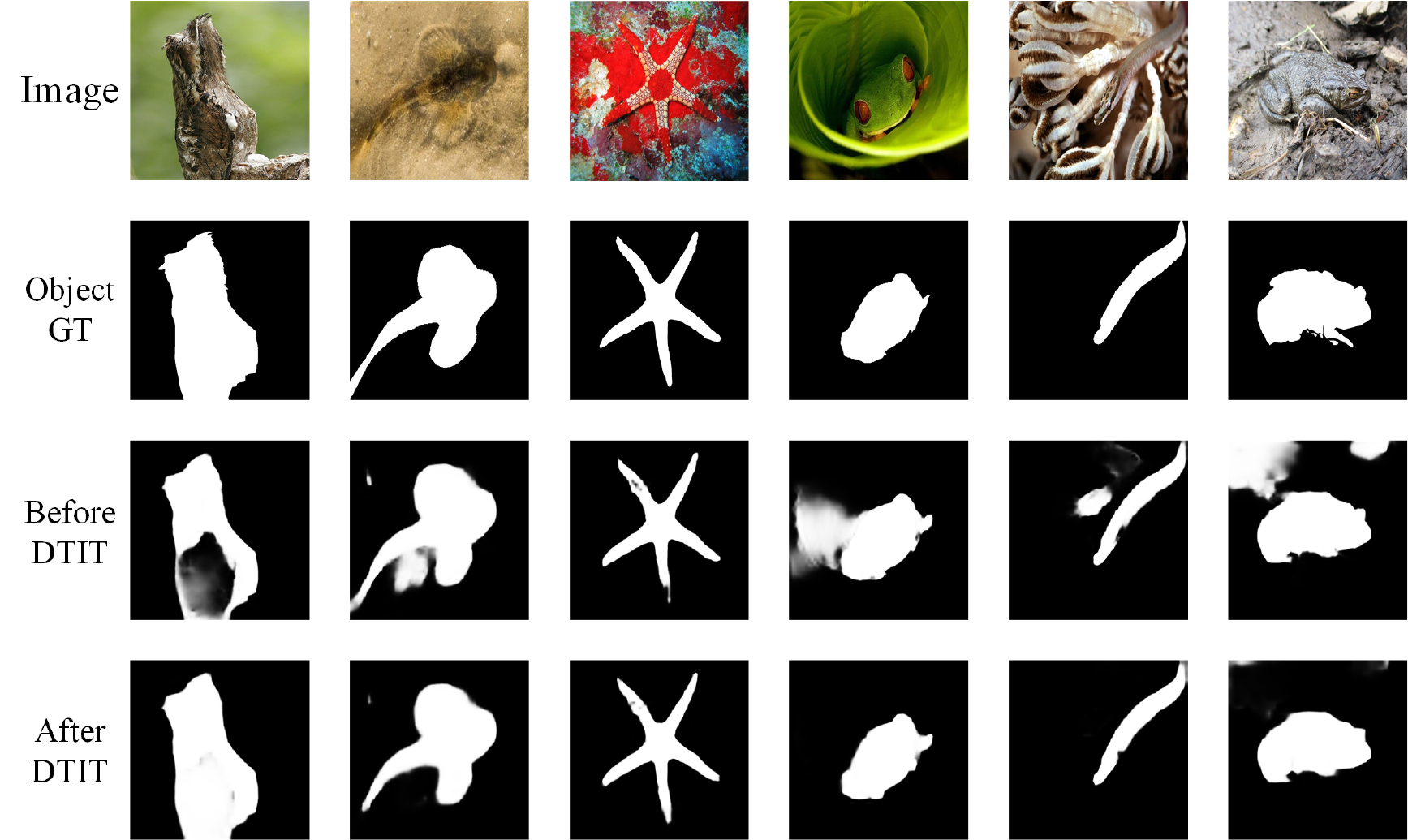} 
\caption{Visual comparison of ablation study about dual-task interactive transformer.}
\label{fig:DTITAblation}
\end{figure}

We also give a visual comparison about the role of DTIT in Fig.\ref{fig:DTITAblation}.
From the top to the bottom, there are input image, camouflaged object ground truth, camouflaged map before DTIT and camouflaged map after DTIT.
We can find that the predicted features after DTIT are purified and the noises are obviously suppressed.

\subsubsection{\textit{Effectiveness of Boundary Generation}}
\begin{table}[!htp]
    \centering
    \setlength{\tabcolsep}{1.0mm}{
    \caption{Ablation experiments of Boundary Generation. The best result is in bold.}
    \label{tab:BoundaryGenerationAblation}
    \resizebox{0.5\textwidth}{!}
{
\begin{tabular}{c|cccc|cccc|cccc}
    \hline\toprule
   \multirow{2}{*}{\centering Variant} & \multicolumn{4}{c|}{\centering CAMO} & \multicolumn{4}{c|}{\centering COD10K} & \multicolumn{4}{c}{\centering NC4K}\\
     &$S_\alpha \uparrow$ & $E_\phi \uparrow$ &$F^\omega_\beta\uparrow$ & MAE$\downarrow$
     &$S_\alpha \uparrow$ & $E_\phi \uparrow$ &$F^\omega_\beta\uparrow$ & MAE$\downarrow$
     &$S_\alpha \uparrow$ & $E_\phi \uparrow$ &$F^\omega_\beta\uparrow$ & MAE$\downarrow$ \\
    \hline
    BoundaryEncoding
        &.852   &.908   & .784  &.053
        &.818   &.889   & .685  &.035
        &.859   &.910   & .784  &.043 \\
    Ours
        &\textbf{.857}&\textbf{.916}&\textbf{.796}&\textbf{.050}
        &\textbf{.824}&\textbf{.896}&\textbf{.695}&\textbf{.034}
        &\textbf{.863}&\textbf{.917}&\textbf{.792}&\textbf{.041} \\
    \bottomrule
    \hline
\end{tabular}}}
\end{table}
Table.~\ref{tab:BoundaryGenerationAblation} shows the effectiveness of boundary generation strategy. ``BoundaryEncoding" adopts a backbone for object encoding and another backbone for boundary encoding.
In our model, boundary is generated from the minus of foreground feature and background feature which are supervised by $GT$ and 1-$GT$ in the highest layer, respectively.
From the comparison, we find that our model shifts uncertain part to boundary, achieving the better boundary extraction.

\begin{figure}[htp!]
\centering
\includegraphics[width=\linewidth]{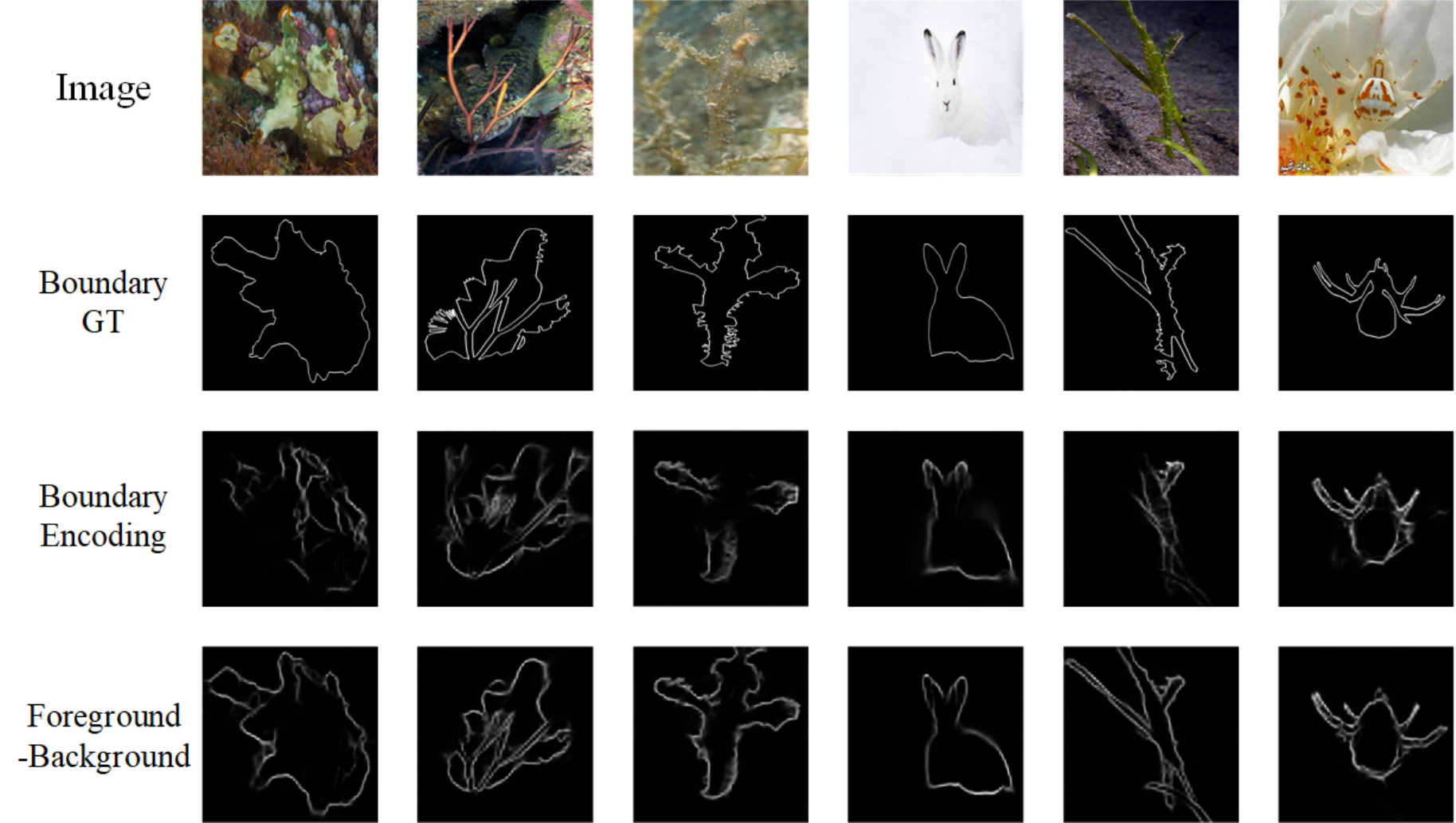} 
\caption{Visual comparison of ablation study about boundary generation.}
\label{fig:BoundaryAblation}
\end{figure}

We also give a visual comparison about two kinds of boundary generation strategies in Fig.\ref{fig:BoundaryAblation}.
From the top to the bottom, there are input image, boundary ground truth, the boundary map obtained from one boundary encoding, the boundary map from our minus operation.
We can observe that our method is consistently excellent in complex scenes ($1^{st}_{}$-$2^{nd}_{}$ columns), similar foreground and background ($3^{rd}_{}$-$4^{th}_{}$ columns) and detail detection ($5^{th}_{}$-$6^{th}_{}$ columns).
Our boundary is located more accurately in $1^{st}_{}$-$2^{nd}_{}$ columns.
Some ignored boundaries in similar foreground and background environment are detected by ours in $3^{rd}_{}$-$4^{th}_{}$ columns. 
More detailed information is provided by ours in $5^{th}_{}$-$6^{th}_{}$ columns, such as the spider's legs.

\section{Conclusion}
We propose to detect camouflaged object and boundary in parallel by cross multi-head self-attention. It fully utilizes the fast and parallel computation power of computer, and surpasses existing COD models which only mimic human vision system.
The proposed dual-task interactive transformer interacts camouflaged object detection task and boundary detection task in the global range, improving the performance of COD.
Besides, the proposed boundary generation strategy
converts blurry boundary region of foreground and background  to sharp boundary by minus operation between foreground and background feature. Supervised by foreground, background and boundary ground truth, our model outperforms state-of-the-art COD models. Nevertheless, our model also encounters high computation cost which will be improved in the future.

\section*{Acknowledgements}
This study is funded by Natural Science Foundation of Anhui Province (1908085MF182).

\bibliographystyle{elsarticle-num}
\bibliography{COD}
\end{document}